\documentclass[11pt]{article}
\usepackage{coling2020}
\usepackage{times}
\usepackage{url}
\usepackage{latexsym}

\usepackage{microtype}
\usepackage{graphicx}
\usepackage{subcaption}
\usepackage{booktabs} 
\usepackage{mathtools}
\usepackage{amsfonts}
\usepackage{amssymb}
\usepackage{amsmath}
\usepackage{multirow}
\usepackage{nicefrac}
\let\vec\boldsymbol
\let\mat\mathbf
\let\ul\underline
\newcommand\F{\mathcal{F}(\vec{x}, \mathcal{W})}
\makeatletter
\newcommand{\ssymbol}[1]{$^{\@fnsymbol{#1}}$}
\makeatother

\newcommand{\citet}[1]{\newcite{#1}}
\colingfinalcopy 

\title{Rethinking Skip Connection with Layer Normalization \\ in Transformers and ResNets}

\author{%
  Fenglin Liu\textsuperscript{1}\thanks{\ \ Equal contribution.}, Xuancheng Ren\textsuperscript{2}\footnotemark[1], Zhiyuan Zhang\textsuperscript{2}, Xu Sun\textsuperscript{2}, Yuexian Zou\textsuperscript{1}\\
  \textsuperscript{1}ADSPLAB, School of ECE, Peking University, China \\
  \textsuperscript{2}MOE Key Laboratory of Computational Linguistics, School of EECS, Peking University \\
  {\tt \{fenglinliu98, renxc, zzy1210, xusun, zouyx\}@pku.edu.cn}
}

\date{}

\begin{document}
\maketitle

\begin{abstract}
Skip connection, is a widely-used technique to improve the performance and the convergence of deep neural networks, which is believed to relieve the difficulty in optimization due to non-linearity by propagating a linear component through the neural network layers. However, from another point of view, it can also be seen as a modulating mechanism between the input and the output, with the input scaled by a pre-defined value one. In this work, we investigate how the scale factors in the effectiveness of the skip connection and reveal that a trivial adjustment of the scale will lead to \textit{spurious} gradient exploding or vanishing in line with the deepness of the models, which could be addressed by normalization, in particular, layer normalization, which induces consistent improvements over the plain skip connection. Inspired by the findings, we further propose to adaptively adjust the scale of the input by recursively applying skip connection with layer normalization, which promotes the performance substantially and generalizes well across diverse tasks including both machine translation and image classification datasets.

\end{abstract}

\section{Introduction}

\blfootnote{
    %
    %
    %
    %
    %
    \hspace{-0.65cm}  
    \quad \ \ This work is licensed under a Creative Commons 
    Attribution 4.0 International License.
    License details:
    \url{http://creativecommons.org/licenses/by/4.0/}.
}

There is a surge of research interest in solving the optimization problem of deep neural networks, which presents a considerable number of novel difficulties that are  in need of practical solutions.
Batch Normalization \cite{Ioffe2015batchnorm}, Layer Normalization \cite{Ba2016layernorm} and Skip Connection \cite{He2016resnetv1} are widely-used techniques to facilitate the optimization of deep neural networks, which prove to be effective in multiple contexts \cite{Szegedy2016Rethinking,Vaswani2017transformer}.

Among them, Batch Normalization (BN) and Layer Normalization (LN) make progress by normalizing and rescaling the intermediate hidden representations to deal with the so-called internal co-variate shift in mini-batch training and collective optimization of multiple correlated features.  
Skip Connection bypasses the gradient exploding or vanishing problem and tries to solve the model optimization problem from the perspective of information transfer. It enables the delivery and integration of information by adding an identity mapping from the input of the neural network to the output, which may ease the optimization and allow the error signal to pass through the non-linearities.

\begin{figure}[t]

    \centering
    \includegraphics[width=1\linewidth]{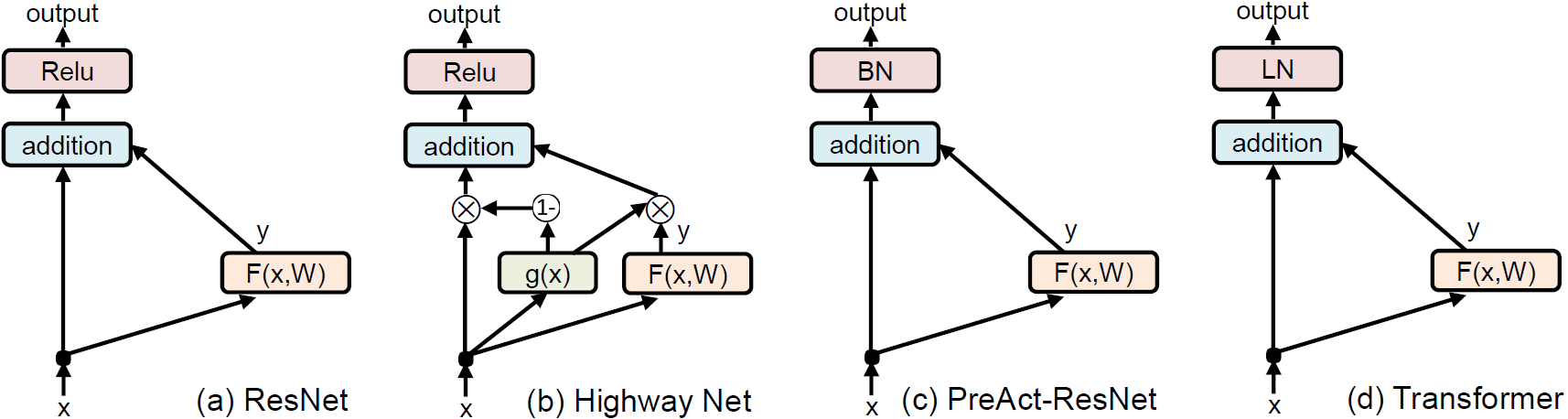}

    \caption{Illustrations of commonly-used skip connections: (a) Conventional Residual Unit \cite{He2016resnetv1}; (b) Residual Unit in Highway Net \cite{Rupesh2015highway}; (c) Residual Unit in PreAct-ResNet \cite{He2016pre-activate-resnet}, where BN represents batch normalization \cite{Ioffe2015batchnorm}; (d) Residual Unit in Transformer \cite{Vaswani2017transformer}, where LN represents layer normalization \cite{Ba2016layernorm}.}
    \label{fig:Residual_Unit}
\end{figure}

Recently, several efforts have been made to deal with the optimization problem by further exploiting the existing solutions, such as exploring the implementation of skip connection \cite{Rupesh2015highway} and combining the skip connection with other normalization methods \cite{He2016pre-activate-resnet,Vaswani2017transformer}. Some of the proposals are illustrated in Figure~\ref{fig:Residual_Unit}. Most of these variations could be formulated as or coarsely equivalent to
\begin{align}
    \vec{y} = \mathcal{G}(\lambda \vec{x} + \F), \label{eq:basic}
\end{align}
where $\vec{x}$ denotes the input of the block, i.e., the skip connection or the shortcut, $\mathcal{F}$ denotes the non-linear transformation induced by the neural network parameterized by $\mat{W}$, $\mathcal{G}$ denotes the normalization function, $\vec{y}$ denotes the output of the residual block, and $\lambda$ denotes the modulating factor that controls the relative importance of the skip connection or the shortcut.\footnote{The terms \textit{skip connection} and \textit{shortcut} are used interchangeably in this work.} A residual model consists of a stack of such blocks. 

A missing piece from the existing work is how would the residual block perform if $\mathcal{G}$ is realized as layer normalization and $\lambda$ does not equal one. Although Transformer \cite{Vaswani2017transformer} has shown the effectiveness of the combination of layer normalization and skip connection, it is intuitive that the modulating factor $\lambda$ may not always be one, especially when $\mathcal{F}$ is not well trained and not sufficient in representation ability.  
In addition, the intuition is supported by \citet{Rupesh2015highway} who found that if using a gate mechanism to determine the balance between the two components, the skip component wins in almost all situations. However, in the study of \citet{He2016pre-activate-resnet}\footnote{They move the batch normalization applied to $\F$ forward and apply it to $\vec{x}$ instead. From another point of view, it is similar to applying the batch normalization to $\vec{y}$ of the previous block.}, which discussed the combination of batch normalization, learned $\lambda$, and convolutional neural networks, they found empirically that the best performance is achieved when $\lambda$ is fixed as one, suggesting the shortcut and the transformed input should be of equal importance. 

\begin{figure*}[t]

    \centering
    \includegraphics[width=1\linewidth]{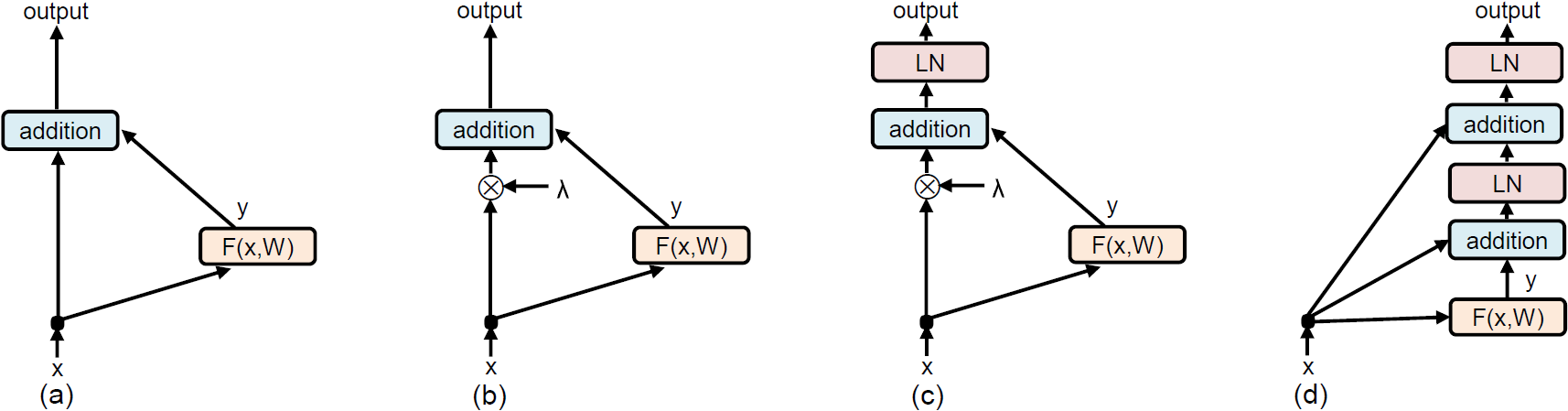}

    \caption{Various combination of skip connection and layer normalization: (a) Original skip connection, (b) Expanded skip connection, (c)  Expanded skip connection with layer normalization, and (d) Recursive skip connection with layer normalization ($\lambda=2$ as an example), where LN represents layer normalization.}
    \label{fig:approach}
\end{figure*}

Therefore, in order to explore whether the optimal ratio of $\vec{x}$ to $\F$ is one in the residual block, and to extend the understandings of the residual block formed as $\vec{y} = \mathcal{G}(\lambda \vec{x} + \F)$,
we propose three different implementations of the residual block in combination with layer normalization.
The first implementation, which we name \textit{Expanded Skip Connection}, simply expands the skip connection and sets $\lambda$ larger than one, following the findings of \citet{Rupesh2015highway}. The second variation, named as \textit{Expanded Skip Connection with Layer Normalization},  includes the layer normalization after the expanded skip connection, since layer normalization is observed to be helpful in facilitating the optimization of skip connection as in \citet{Vaswani2017transformer}. The final proposal, \textit{Recursive Skip Connection with Layer Normalization}, is a novel combination that does not fit in the general form of the residual block, which takes the advantages of skip connection and layer normalization in a recursive manner, so that the input signal is amplified yet with layer normalization stabilizing the optimization. The illustrations of those combinations are shown in Figure~\ref{fig:approach}. 
Our experiments on diverse tasks, including machine translation and image classification with strong baselines, i.e., Transformer and ResNet, conclude the following:
\begin{itemize}
    \item The expanded skip connection without any normalization indeed causes serve gradient malformation, leading to unsatisfactory learning of the neural networks. Layer normalization helps to deal with the optimization problem introduced by the expanded skip connection to a certain extent.
    \item The proposed recursive skip connection with layer normalization further facilities the optimization by separating the expanded skip connection into multiple stages to better incorporate the effect of the transformed input. It achieves comfortable margin on datasets from different domains in extensive experiments to determine the proper granularity of the skip connection. 
    \item The experimental results on the WMT-2014 EN-DE machine translation dataset using Transformer further proves the effectiveness and the efficiency of the recursive architecture, which helps a model to perform even better than the model three times larger.
\end{itemize}

\section{Related Work}
In recent years, a large number of skip connection methods have been used for neural networks \cite{He2016resnetv1,Rupesh2015highway,He2016pre-activate-resnet,Vaswani2017transformer,Szegedy2017Connections,Wang2019Learning,Nguyen2019Transformers,Liu2020CVD,Xu2019Understanding,Bachlechner2020ReZero,Liu2020Understanding,Shen2020PowerNorm,Xiong2020LNTransformer}. For the skip connection, there are two main types of problems. The first is how should the skip connection be incorporated into the existing neural network structures so that the best improvements can be achieved. The second is how should the neural network with skip connections be optimized so that its representation capability could be fully mined. Therefore,  we categorized the related work into \textit{On the Connection Problem} and \textit{On the Optimization Problem}.

\paragraph{On the connection problem}
Highway network~\cite{Rupesh2015highway} built a highway connection from the input to the output, similar to the skip connection. However, a transform gate was proposed to control the balance of the input $\vec{x}$ and the transformed input $\F$, instead of using identity mapping to combine the input and the transformed input.
\citet{He2016pre-activate-resnet} designed five types of skip connections and discussed the possible skip connections in detail. Based on their theory and experiments, they argued that it is likely for a model to perform the best when the skip connection is directly combined into the output without any additional scaling such that the skip and the transformed input have equal contribution to the output. The reason is that with scaling, the gradient of the skip suffers from the gradient exploding or vanishing problem, which hinders the deep neural network from efficient optimization.

\paragraph{On the optimization problem}
Since the early work has established that it is sufficient to directly combine skip connection  into forward propagation of neural networks without any scaling \cite{He2016pre-activate-resnet}, the succeeding studies on the optimization problem mostly followed the simple yet effective architecture of the skip connection. In computer vision task, PreAct-ResNet~\cite{He2016pre-activate-resnet} demonstrated that it is helpful to move the batch normalization applied to $\F$ forward and apply it to $\vec{x}$ instead. From another point of view, it is similar to applying the batch normalization to $\vec{y}$ of the previous block.
In natural language processing tasks, Transformer~\cite{Vaswani2017transformer} made use of skip connection and layer normalization extensively in its architecture, and from our experiments, the layer normalization seems essential to the overall performance of the model and can help the optimization of the non-linear transformation to some extent.

In all, we propose to use the form $\vec{y} = \mathcal{G}(\lambda\vec{x} + \F)$ to summarize the combination of normalization and skip connection in existing work. The form is in accordance with the findings that the normalization $\mathcal{G}$ should be placed after the addition of the skip and it could be beneficial to elevate the portion of the input information with $\lambda$. Moreover, this work distinguishes itself in that (a) we mainly focus on the combination of layer normalization and skip connection, which is less investigated but deems promising, (b) we rethink the common practice that the skip need not scale in consideration of the help of layer normalization, (c) we conduct experiments on both representative computer vision and natural language processing tasks, while existing work only provides result in a single modality, and most importantly (d) we escape from the general form of the residual block that generalizes all previous work and propose a novel recursive construction of the residual block with layer normalization that outperforms all the variants of the general form that are examined in this work.

\section{Architecture}\label{sec:arch}
In this section, we describe three skip connection constructions, which are illustrated in Figure~\ref{fig:approach}. They are based on the integration of more input information and application of the layer normalization. 

\paragraph{E\emph{x}panded \emph{Skip} Connection (xSkip)}

The expanded skip connection is defined similar to the constant scaling method in \citet{He2016pre-activate-resnet} as
\begin{equation}
    \vec{y} = \lambda \vec{x} + \F,
\end{equation}
where $\vec{x}$ and $\vec{y}$ are the input and the output of the residual block, respectively. $\mathcal{F}$ denotes the weighted neural network layer, and $\lambda$ is a modulating scalar. Naturally, this construction adjusts the importance of the skip considering that the neural network layer may come with different representation ability and optimization difficulty. However, it should be noted that in this work $\lambda$ is fixed to isolate the effect of scaling. While it is possible that a learned $\lambda$ may better capture the balance between the two components, the learning of $\lambda$ becomes another variable. Following \citet{Rupesh2015highway}, we mainly experiment with $\lambda>1$.

\paragraph{E\emph{x}panded \emph{Skip} Connection with Layer Normalization (xSkip+LN)}

Motivated by Transformer, which combines skip connection with layer normalization, we further examine the effect of layer normalization on expanded skip connection, which takes the form of
\begin{equation}
    \vec{y} = \text{LN}(\lambda \vec{x} + \F).
\end{equation}
Layer normalization may help mitigate the gradient malformation in optimization caused by the modulating factor. We use the parameterized layer normalization as suggested in \citet{Ba2016layernorm}, because different input features naturally have different importance.
Different from batch normalization which acts in ``sample space'', layer normalization acts in ``feature space''.
Considering the special case that $\F$ is ill-learned and outputs random values $\vec{\epsilon}$ and the examples $\mat{X} = \{\vec{x}_i\}$ in the mini-batch all contain the same input feature $f$ of the same value $v$, batch normalization simply ignores the input feature $f$ which results in uneventful optimization according to mini-batch statistics, while layer normalization still recognizes the input feature $f$ and its value can be properly adjusted according to the other input features. The same may also apply to the situation where $\lambda\vec{x}$ is dominant compared to $\F$ and the transformed input falls into the range comparable to random noise. In such situations, which may happen when the neural network is hard to optimize, layer normalization could still help the learning of the shortcut, while batch normalization might fail.

\paragraph{\emph{R}ecursive \emph{Skip} Connection with \emph{L}ayer \emph{N}ormalization (rSkip+LN)}

Another way to stabilize the gradient is to keep $\lambda=1$ each time but repeatedly add the shortcut with layer normalization, such that more input information is also modeled. It is defined recursively as
\begin{align}
    \vec{y}_\lambda = \text{LN}(\vec{x} + \vec{y}_{\lambda-1})\quad \text{and} \quad 
    \vec{y}_1       = \text{LN}(\vec{x} + \F),
\end{align}
where $\lambda$ should be an integer no smaller than one. For example, when $\lambda=1$, it regresses to the block used in Transformer and conforms to the findings that the skip need not scale. A further example is when $\lambda=2$, $\vec{y}_2 = \text{LN}(\vec{x} + \vec{y}_1) = \text{LN}(\vec{x} + \text{LN}(\vec{x} + \F))$, which is shown in Figure~\ref{fig:approach}. Through recursive skip connection with layer normalization, the model is encouraged to use layer normalization multiple times to improve optimization, and can incorporate more information of $\vec{x}$ by the skip connection. Moreover, the model may gain more expressive power than simply incorporating the scaled skip at once, since each recursive step essentially constructs a different feature distribution and the recursive structure can learn the adaptive ratio of $\vec{x}$ to $\F$.
The ratio between the input $\vec{x}$ and the residual $\F$ is
    $\nicefrac{\vec{x}}{\F} = \nicefrac{\sigma_1}{\vec{w}_1} + 1$, where $\sigma_1$ is the standard deviation of $\vec{x}+\F$ and $\vec{w}_1$ is the gain parameters of the inner layer normalization, which penetrate through the normalization and effectively act as the learnable ratio.
It is also important to note that the ratio is also dependent on the current layer distribution, which could serve as a regularization component. For any $\lambda$ larger than 2, the general form of the ratio is $1 + \sum_{i=1}^{\lambda} \prod_{j=1}^{i} \nicefrac{\sigma_j}{\vec{w}_j}$. 
This derivation, which is detailed in Appendix~\ref{app:ratio}, also indicates that setting $\lambda=2$ should be sufficient for learning the shortcut-residual ratio and adjusting the ratio based on instant data distribution. Yet with larger $\lambda$, the distribution of the higher order combination of $\vec{x}$ and $\mathcal{F}(\vec{x})$ will be considered. The recursive definition introduces gradual and smooth combination of the input and the residual up to the specified order $\lambda$ which could deal with different distributions better.

\section{Empirical Study}

In this section, we examine the effect of the discussed combinations on two representative baselines through empirical studies. First, we conduct exploratory experiments on datasets that are relatively small to gain insights of the combinations, such as the choice of $\lambda$ and the effectiveness of the constructions. Then, we discuss the observed results and conduct further analysis. Finally, we validate our findings on larger and more difficult datasets.

\subsection{Tasks and Settings}

For ease of introduction, brief descriptions of the tasks and corresponding models are summarized together in the following. As the combinations to be examined only relates to the skip connection and we keep the inner structure of the residual block untouched, please refer to the cited work for more detailed description of the baselines.

\subsubsection{Image Classification}

\paragraph{Datasets}

CIFAR-10 and CIFAR-100~\cite{Krizhevsky2009cifar10} are two datasets for colored image classification. Both contain 50K images for training and 10K images for testing. The former has 10 classes, while the latter has 100 classes and is considered more difficult. We report the classification error rates as the evaluation of the performance.

\paragraph{Baselines and settings}

We conduct experiments using the ResNet with pre-activation (PreAct-ResNet) as proposed in \citet{He2016pre-activate-resnet}. For experiments on CIFAR-10, we use ResNet-110 for exploration because of its deep structures. It contains 54 two-layer residual blocks, which makes it a challenging baseline for optimization. We validate the proposal on CIFAR-100, using a variety of ResNet structures, such as ResNet-20, ResNet-32, ResNet-44, ResNet-56, and ResNet-110, to ensure the generality of the results.
We mostly adopt the hyperparameters from \citet{He2016pre-activate-resnet} to keep the number of variables controlled and comparable, except that we use a weight decay rate of 0.0002.

\subsubsection{Machine Translation}

\paragraph{Datasets}

IWSLT 2015 English-Vietnamese dataset (EN-VI)  \cite{Cettolo2015IWSLT} consisting of 133K sentence pairs, WMT 2014 English-German  dataset (EN-DE) consisting of 4.5M sentence pairs and WMT 2014 English-French dataset (EN-FR) consisting of 36M sentence pairs are for machine translation. Following previous work \cite{Luong2015Stanford}, we use \texttt{tst2012} and \texttt{tst2013} as development set and test set, respectively for EN-VI dataset. For EN-DE, we use newstest2013 for development and newstest2014 for testing.
For EN-FR, we validate on newstest2012+2013 and test on newstest2014.
The evaluation metric is BLEU~\cite{Papineni2002bleu} against the reference translations.

\paragraph{Baselines and settings}

Transformer~\cite{Vaswani2017transformer} is a representative model for machine translation, which includes skip connection and layer normalization in the model structure. Transformer follows the encoder-decoder framework, except that it substitutes the conventional recurrent layers with the self-attention based blocks, i.e., $\F$ consists of a dot-product based attention layer and two feed-forward layers. Although Transformer is not as deep as ResNet-110, it is heavily parameterized and the commonly-used six-layer Transformer-Base structure embodies 65M learnable parameters, which also poses challenge for optimization.
We base our experiments on the tensor2tensor \cite{Vaswani2018Tensor2Tensor} code. For experiments on IWSLT-2015, we set batch size to 4096 and train on single GPU, as it is relatively small. For experiments on WMT-2014 that is much larger, due to the complication caused by the special batching algorithm and the distributed learning, we find setting batch size to 16,384 tokens in total on 3 GPUs can reproduce the reported results on 8 GPUs using batch size of 32,768 tokens in total \cite{Vaswani2017transformer}. However, the training steps required are increased from 100,000 steps to 550,000 steps. The rest of hyperparameters are the same as \citet{Vaswani2017transformer}.

\begin{table*}[t]
    \centering
    \footnotesize
    \vskip 0.1in
    \begin{tabular}{@{}l l c c c@{}}
        \toprule
        Method   & Architecture                                                  & $\lambda$ & $\mathcal{G}$ & Error (\%)    \\ \midrule
        Baseline & $\vec{x}+\F$                                                  & 1         & -             & 6.37          \\ \midrule
        1xSkip    & $\vec{x}+\F$                                                  & 1         & -             & \ul{6.31}     \\
        2xSkip    & $2\vec{x}+\F$                                                 & 2         & -             & 8.41          \\
        3xSkip    & $3\vec{x}+\F$                                                 & 3         & -             & 9.23          \\ \midrule
        1xSkip+LN & $\text{LN}(\vec{x} + \F)$                                     & 1         & LN            & 7.72          \\
        2xSkip+LN & $\text{LN}(2\vec{x} + \F)$                                    & 2         & LN            & 6.29          \\
        3xSkip+LN & $\text{LN}(3\vec{x} + \F)$                                    & 3         & LN            & \ul{6.07}     \\ \midrule
        1rSkip+LN & $\text{LN}(\vec{x}+\F)$                                       & 1         & LN            & 7.72          \\
        2rSkip+LN & $\text{LN}(\vec{x}+\text{LN}(\vec{x}+\F))$                    & 2         & LN            & \ul{\bf 6.02} \\
        3rSkip+LN & $\text{LN}(\vec{x}+\text{LN}(\vec{x}+\text{LN}(\vec{x}+\F)))$ & 3         & LN            & 6.55          \\
        \bottomrule
    \end{tabular}
    \caption{Results on CIFAR-10 using PreAct-ResNet-110. Average of 5 runs. The lower the better. The underlined numbers denote the best results under the same comparable settings. $\lambda$xSkip, $\lambda$xSkip+LN, and $\lambda$rSkip+LN stand for expanded skip connection, expanded skip connection with LN, and recursive skip connection with LN, respectively. Please note that 1xSkip is equivalent to Baseline and 1rSkip+LN is equivalent to 1xSkip+LN. It is clear that simply expanding the skip connection hurts the performance and using layer normalization can mitigate the problem.}
    \label{tab:c10-res-main}
\end{table*}

\subsection{Exploratory Results}

In this section, we evaluate the effect of the proposed constructions and report the exploratory results on CIFAR-10 using ResNet and IWSLT-2015 using Transformer. 

The main results on CIFAR-10 using PreAct-ResNet-110 are summarized in Table~\ref{tab:c10-res-main}. We experiment with different constructions and different $\lambda$s. 1xSkip is our reimplementation of the ResNet-110 with pre-activation and outperforms the reported result in \citet{He2016pre-activate-resnet} by 0.06. This margin could approximately serve as the boundary deciding whether the improvement is caused by random variations or experimental settings. As we can see, for skip connection without normalization, it is indeed the best to add the input without scaling and expanding the skip results in significant loss in accuracy. However, with the application of layer normalization, the expanded skip can also be sufficiently optimized and surpass the baseline, which is suggested by \citet{He2016pre-activate-resnet} by extensive structure search. Moreover, with the recursive design, the performance can also be maintained and yet further improved, which reaches the best result in our experiments when using $\lambda=2$. It is also interesting to see that in combinations with layer normalization, the $\lambda$ that results in the best accuracy is also different. It is intuitive that the recursive structure adjusts the effective ratio of the skip and the transformed input more sophisticatedly than the plain expanding structure so that different $\lambda$ may be required. 

Table~\ref{tab:envi-transformer-main} shows the results on IWSLT-2015 using Transformer and the results about layer normalization are similar. With the increasing of the weight of the skip and layer normalization, the performance of the model improves. Since the models incorporated expanded skip connection with layer normalization have the same number of parameters, regardless of $\lambda$, thus embodying the same representation ability, the improved performance should mainly be attributed to the enhanced shortcut information. Moreover, the recursive version brings significant improvements upon the simply expanding version, pushing the BLEU score past 31 and reaching a margin of 1.1 BLEU scores at most, demonstrating the advantage of gradually adding the expanded skip to ease the optimization. A notable difference from the results on PreAct-ResNet-110 is that the plain skip connection version does not perform better than the one with layer normalization (30.31 BLEU vs. 28.98 BLEU on Transformer and 92.28 accuracy vs. 93.69 accuracy on PreAct-ResNet-110). We find that the reason is likely to be that in the residual block of PreAct-ResNet-110, there are already two instances of batch normalization before each convolutional layer to facilitate its optimization, while in the Transformer block no other normalization technique exists.

In summary, our experiments show that feeding more input signals to the layer output should be helpful to learning of the overall model. Previous concerns on the optimization difficulty introduced by the expanded skip connection could be alleviated by the usage of layer normalization. Especially, the proposed recursive skip connection with layer normalization brings the most improvements by dividing the expanded skip and iteratively adding the pieces to the output, which can further stabilize the gradient on the input and help the training of the weighted layer.

\begin{table*}[t]
    \centering
    \footnotesize
    \vskip 0.1in
    \begin{tabular}{@{}l l c c c@{}}
        \toprule
        Method & Architecture                                                  & $\lambda$ & $\mathcal{G}$ & BLEU           \\ \midrule
        Baseline    & $\text{LN}(\vec{x}+\F)$                                       & 1         & -             & 30.31          \\ \midrule
        1xSkip       & $\vec{x}+\F$                                                  & 1         & -             & 28.98          \\ \midrule
        1xSkip+LN    & $\text{LN}(\vec{x} + \F)$                                     & 1         & LN            & 30.31          \\
        2xSkip+LN    & $\text{LN}(2\vec{x} + \F)$                                    & 2         & LN            & \ul{30.99}     \\
        3xSkip+LN    & $\text{LN}(3\vec{x} + \F)$                                    & 3         & LN            & 30.92          \\ \midrule
        1rSkip+LN    & $\text{LN}(\vec{x}+\F)$                                       & 1         & LN            & 30.31          \\
        2rSkip+LN    & $\text{LN}(\vec{x}+\text{LN}(\vec{x}+\F))$                    & 2         & LN            & \ul{\bf 31.45} \\
        3rSkip+LN    & $\text{LN}(\vec{x}+\text{LN}(\vec{x}+\text{LN}(\vec{x}+\F)))$ & 3         & LN            & 31.10          \\
        \bottomrule
    \end{tabular}
    \caption{Results on EN-VI machine translation dataset using Transformer. The higher the better. $\lambda$xSkip, $\lambda$xSkip+LN, and $\lambda$rSkip+LN are defined similarly. Please note that Baseline, 1xSkip+LN, and 1rSkip+LN are equivalent. It is clear that without layer normalization, Transformer experiences significantly performance degradation. By expanding the shortcut, substantially better results can be achieved. The best improvements of over 1.1 BLEU scores are brought by the proposed recursive skip connection.}
    \label{tab:envi-transformer-main}
\end{table*}


\subsection{Discussions}

However, questions remain whether the improvements can also be achieved by contracting skip connection and whether batch normalization can achieve the same effect which is more commonly in convolutional neural networks. We conduct the following analysis to answer these questions. Furthermore, we demonstrate that layer normalization adaptively adjusts the gradients of the scaled skip, solving the gradient malformation caused by the modulating factor $\lambda$.

\paragraph{Contracting skip connection}\label{sec:discussion}

We experiment two different ways to relatively tune down the skip ratio, i.e., using smaller $\lambda$ and expanding $\F$. The results are shown in Table~\ref{tab:c10-res-low}. As we can see, if the shortcut is forced to play the minor role, the deep neural network fails to be optimized properly and shows catastrophic performance. In contrast, the expanded skip causes less harm to the overall performance (cf. Table~\ref{tab:c10-res-main}), indicating the importance of linear component in the optimization of deep neural networks. It is consistent with the findings of \citet{Rupesh2015highway} and \citet{He2016pre-activate-resnet}. Moreover, the results also show that applying layer normalization under this situation further deteriorates the performance.

\paragraph{Using batch normalization}

Since batch normalization has been widely used in CNNs, we replace layer normalization with batch normalization to see the difference. However, comparing Table~\ref{tab:c10-res-main} and Table~\ref{tab:c10-res-bn}, it can be seen that batch normalization is less effective than layer normalization in combination with skip connection. The proposed recursive structure can help batch normalization deal with the expanded skip as each time the ratio is still kept as one. As we explained in Section~\ref{sec:arch}, batch normalization makes use of mini-batch statistics which may prevent the plainly expanded skip from coming into effect.

\begin{figure*}[t]

    \centering

    \includegraphics[width=0.24\linewidth]{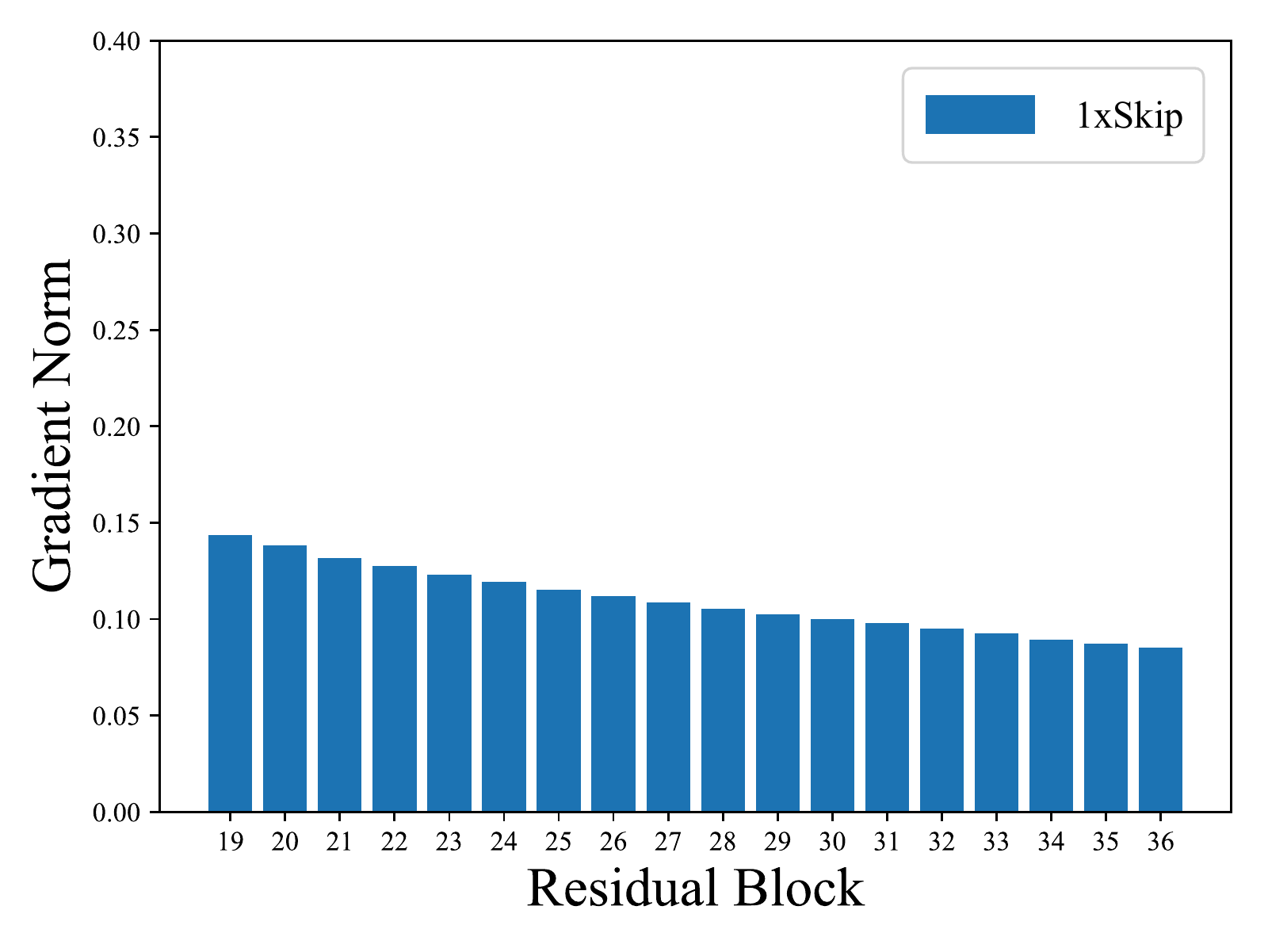}
    \hfill
    \includegraphics[width=0.24\linewidth]{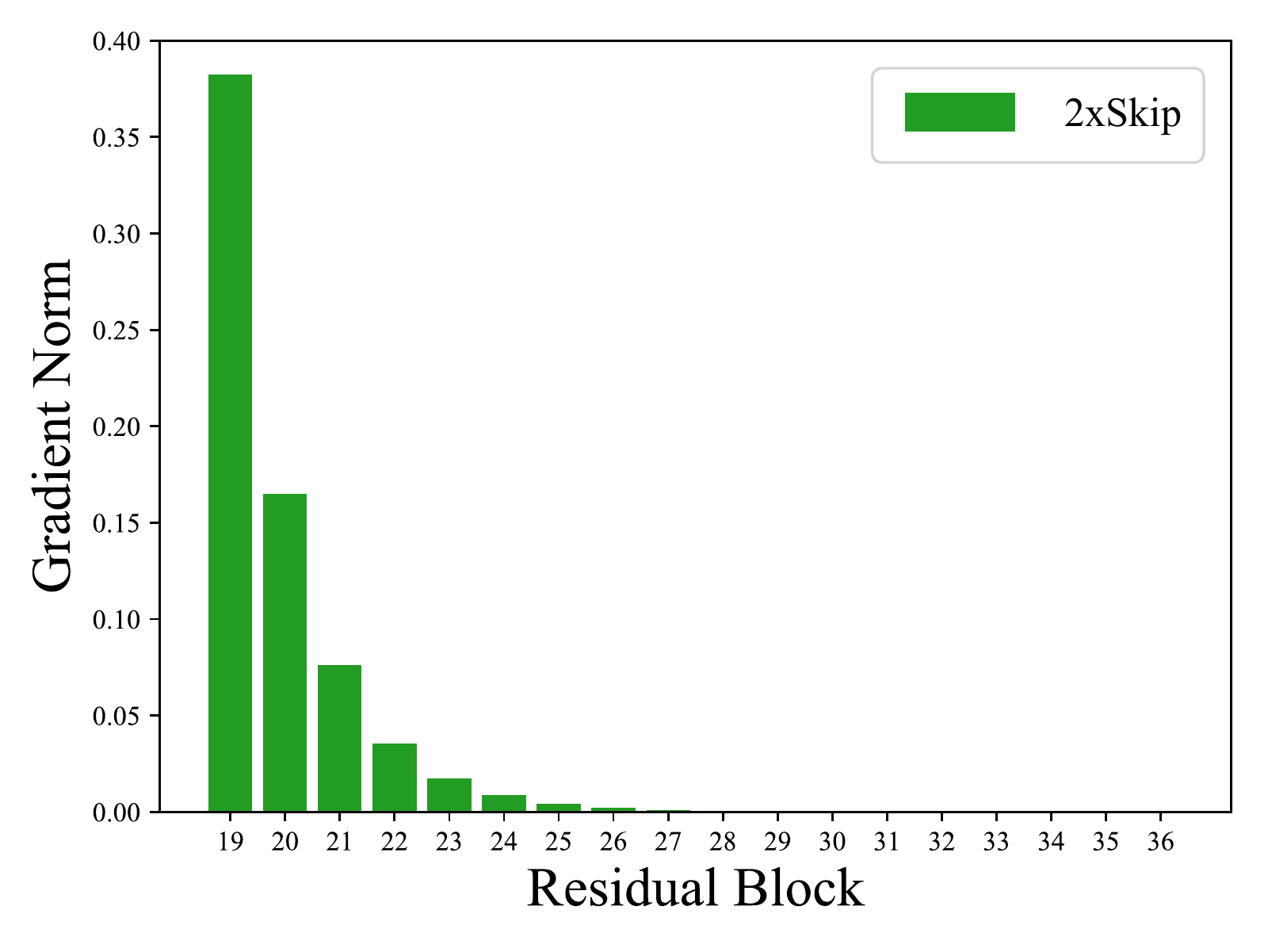}
    \hfill
    \includegraphics[width=0.24\linewidth]{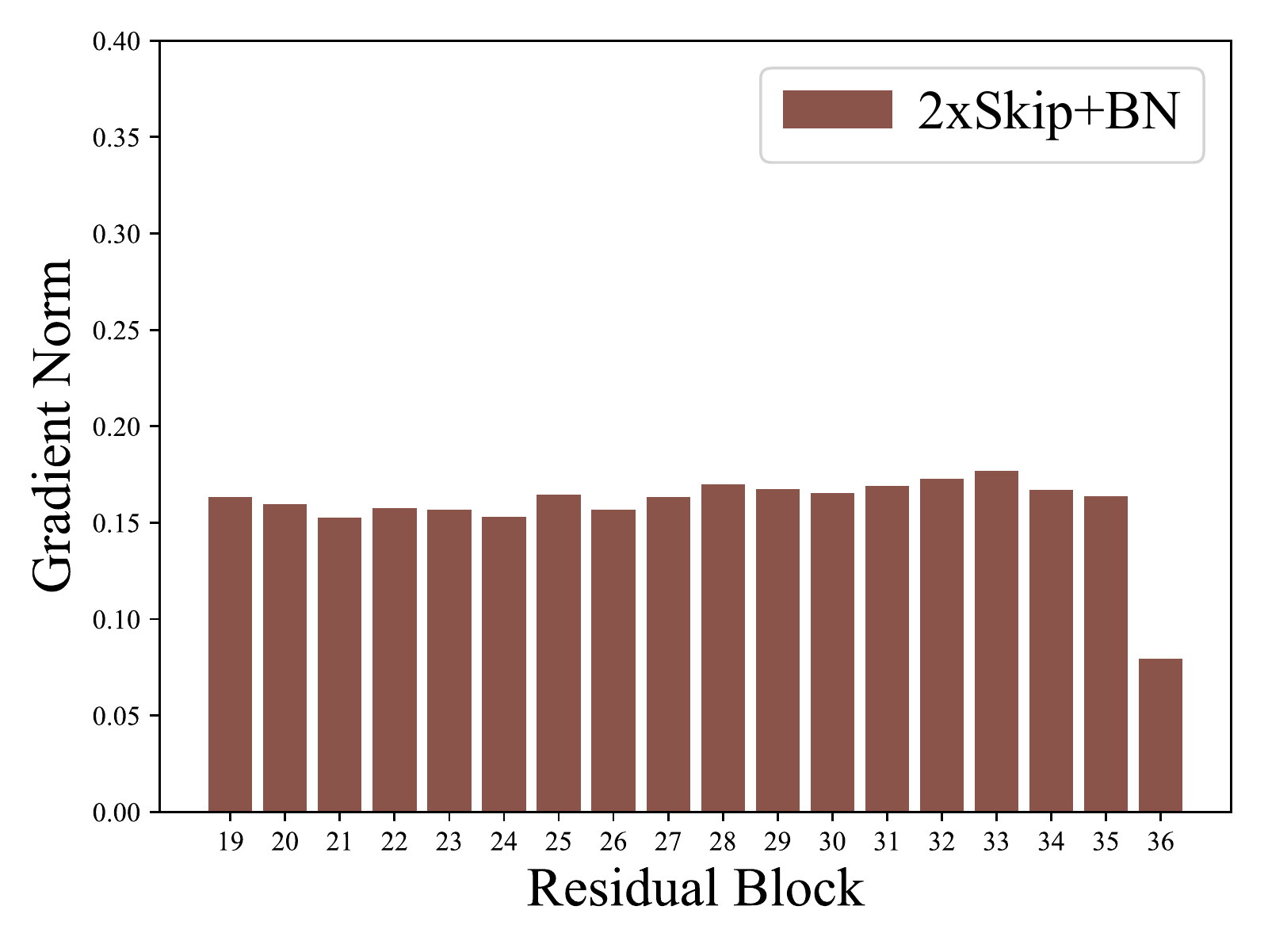}
    \hfill
    \includegraphics[width=0.24\linewidth]{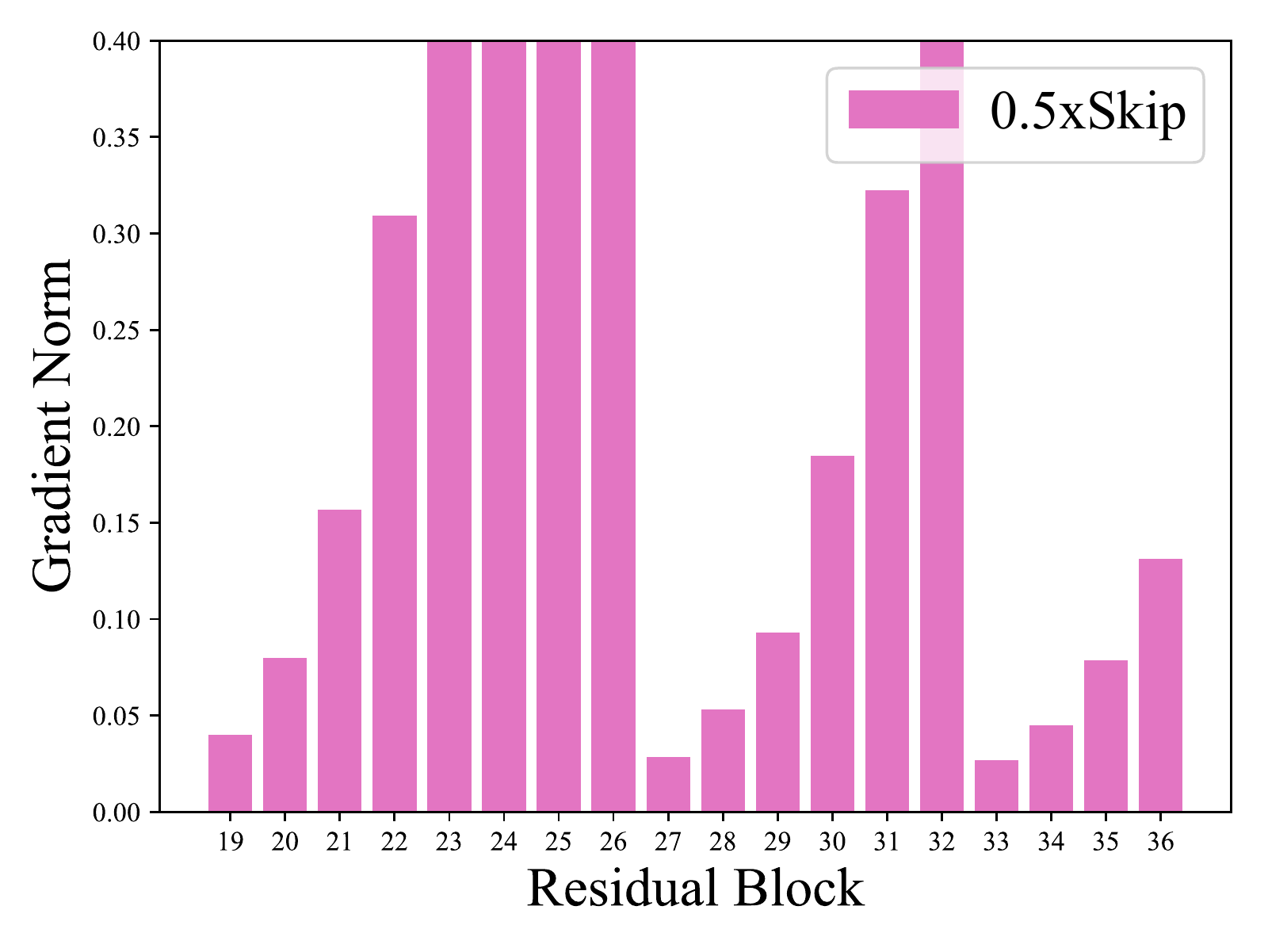}
    
    \includegraphics[width=0.24\linewidth]{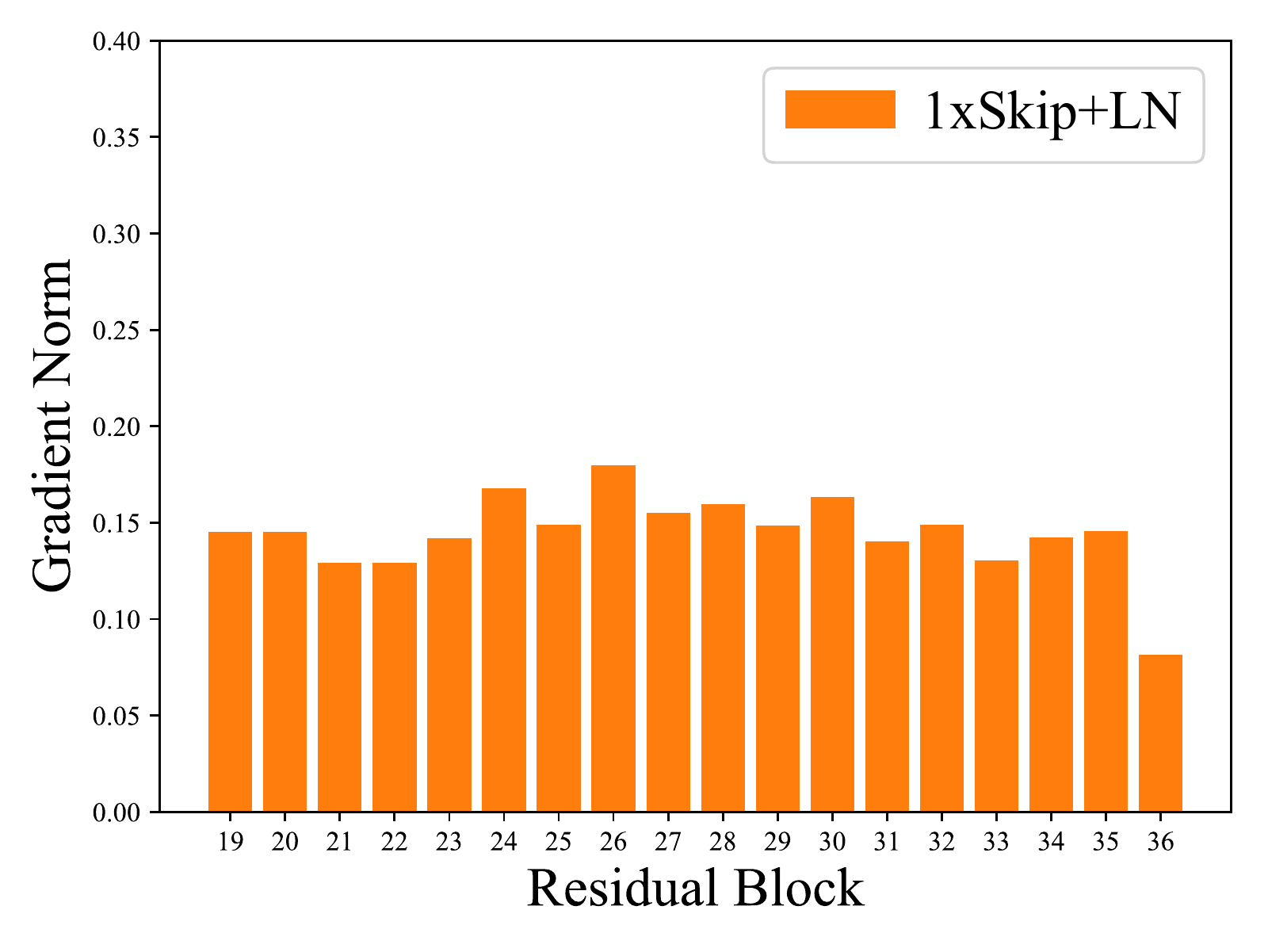}
    \hfill
    \includegraphics[width=0.24\linewidth]{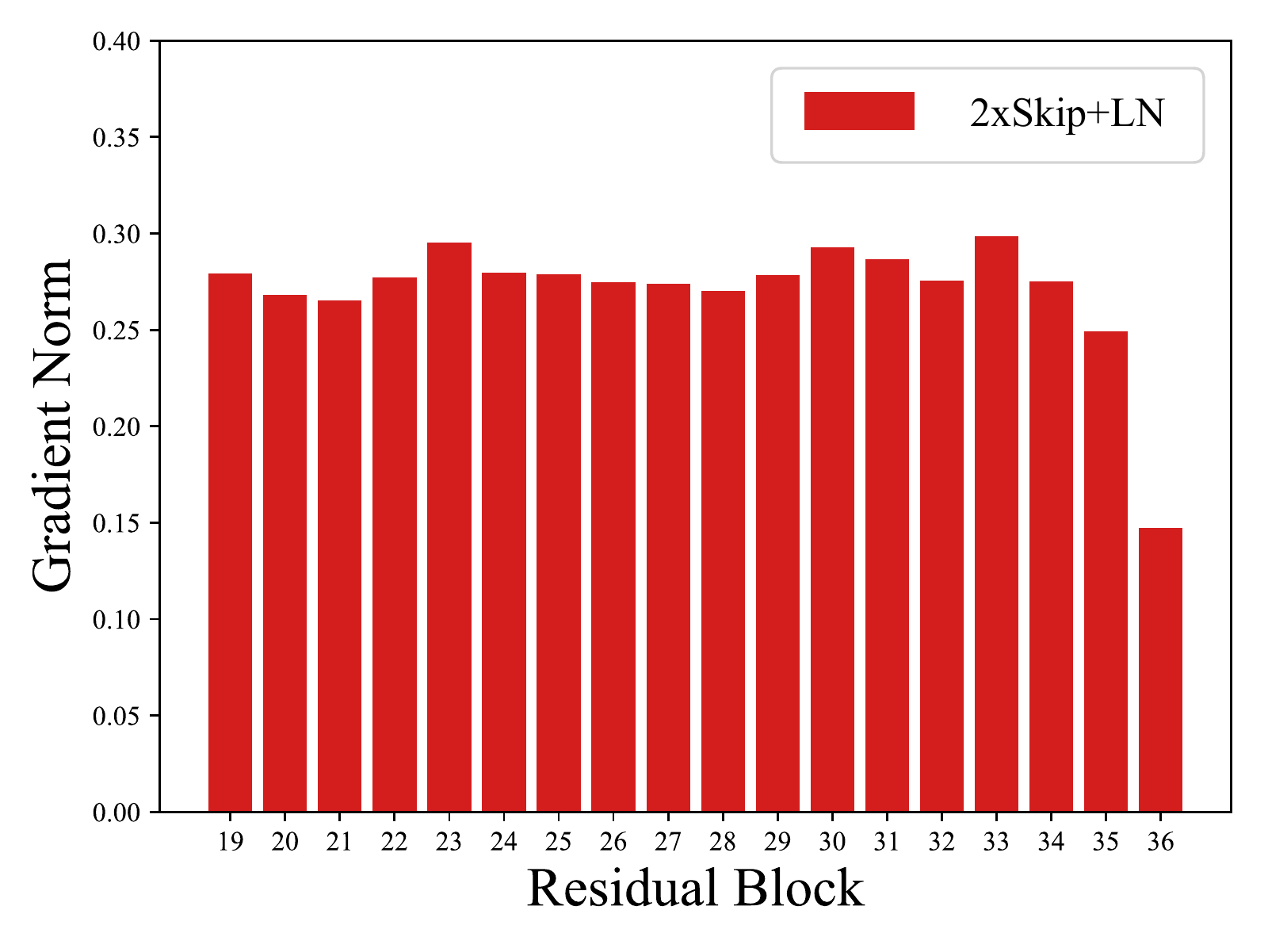}
    \hfill
    \includegraphics[width=0.24\linewidth]{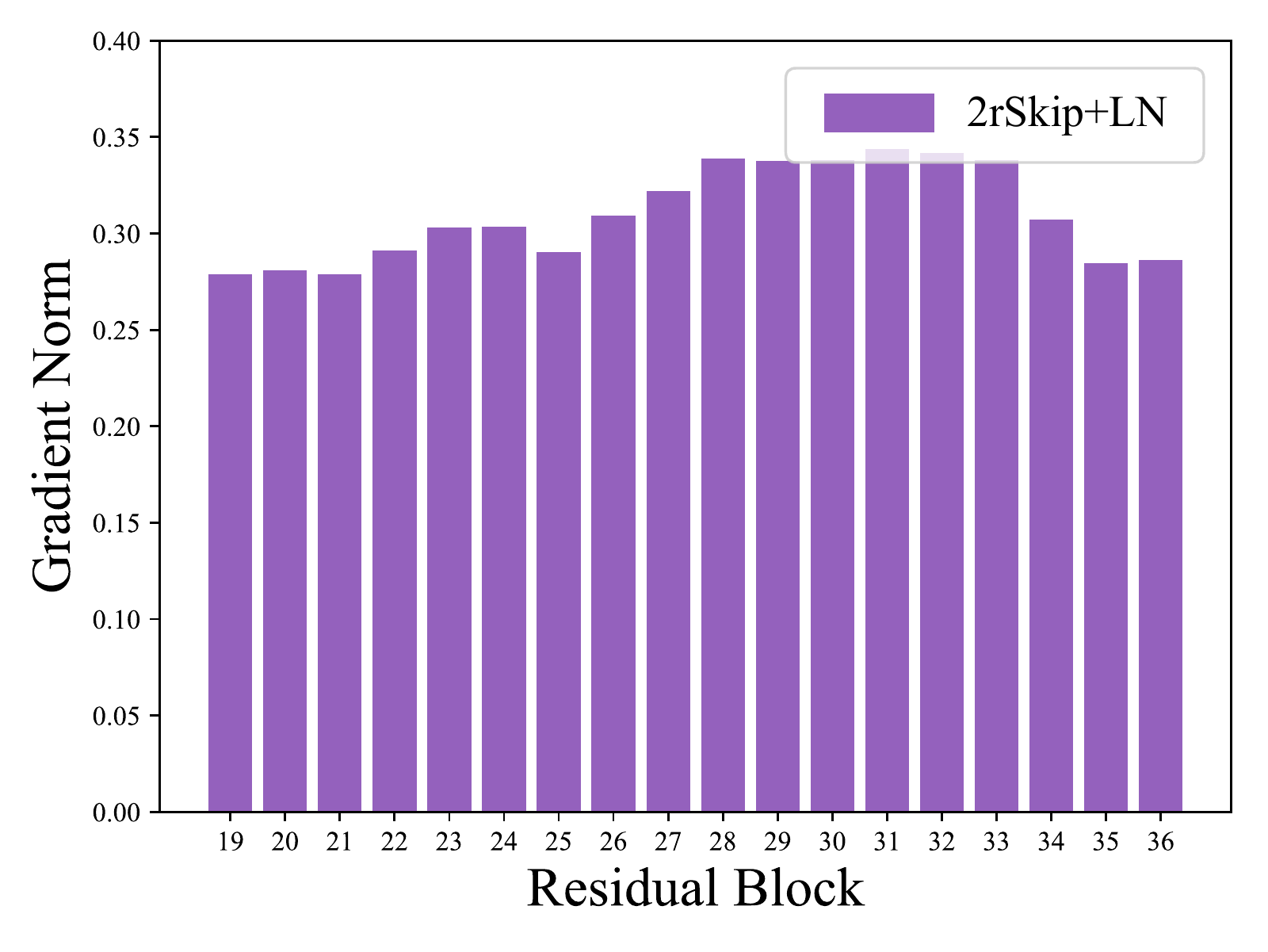}
    \hfill
    \includegraphics[width=0.24\linewidth]{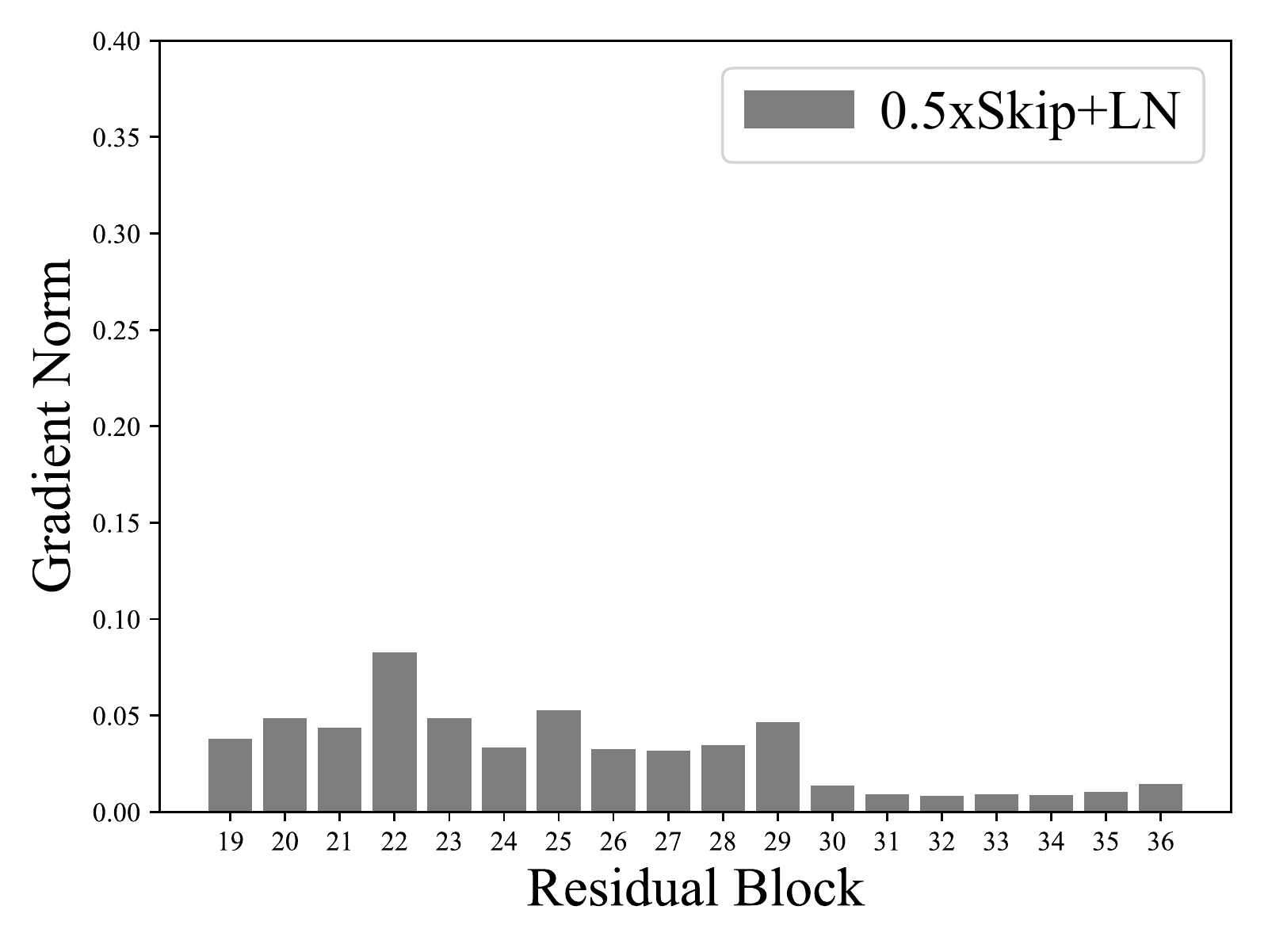}

    \caption{Gradient norm of the output of residual blocks of various constructions on CIFAR-10 using PreAct-ResNet-110. We show the gradient norm of the output $\vec{y}$ of the residual blocks of middle depth in the model averaged over 10,000 random examples. By definition, $\vec{y}$ equals $\vec{x}$ of the following block.}
    \label{fig:grad}
\end{figure*}

\begin{table}[t]
\parbox{.32\linewidth}{
    \centering
    \footnotesize
    \begin{tabular}{@{}l c@{}}
        \toprule
        Architecture               & Error (\%)       \\ \midrule
        $\vec{x}+\F$               & \phantom{0}6.31 \\ \midrule
        $0.5\vec{x}+\F$            & 18.61           \\ \midrule
        $\text{LN}(0.5\vec{x}+\F)$ & 29.88           \\
        $\text{LN}(\vec{x}+2*\F)$  & 24.98           \\
        $\text{LN}(\vec{x}+3*\F)$  & 61.02           \\ \bottomrule
    \end{tabular}
    \caption{Results on CIFAR-10 with contracted skip connection. Lowering the skip ratio leads to catastrophic results and LN further aggravates the performance.}
    \label{tab:c10-res-low}
}
\hfill
\parbox{.32\linewidth}{
    \centering
    \footnotesize
    \begin{tabular}{@{}l c c c@{}}
        \toprule
        ResNet-110 & $\lambda$ & $\mathcal{G}$ & Error (\%) \\ \midrule
        1xSkip      & 1         & -             & 6.31      \\ \midrule
        1xSkip+BN   & 1         & BN            & 8.23      \\
        2xSkip+BN   & 2         & BN            & 6.43      \\
        3xSkip+BN   & 3         & BN            & 6.50      \\ \midrule
        1rSkip+BN   & 1         & BN            & 8.23      \\
        2rSkip+BN   & 2         & BN            & 6.77      \\
        3rSkip+BN   & 3         & BN            & 6.68      \\
        \bottomrule
    \end{tabular}
    \vspace{-0.6\baselineskip}
    \caption{Results on CIFAR-10 with BN. Compared to LN, BN is less effective dealing with the expanded skip connection. }
    \label{tab:c10-res-bn}
}
\hfill
\parbox{.32\linewidth}{
    \centering
    \footnotesize
    \begin{tabular}{@{}l c c c@{}}
        \toprule
        Transformer & \# Blocks & BLEU           \\ \midrule
        1xSkip+LN    & 6         & 30.31          \\
        2rSkip+LN    & 6         & \ul{\bf 31.45} \\ \midrule
        1xSkip+LN    & 2         & 30.04          \\
        2rSkip+LN    & 2         & \ul{30.49}     \\
        \bottomrule
    \end{tabular}
    \vspace{0.2\baselineskip}
    \caption{Results on Transformer of fewer blocks. As we can see, with the proposed combination of recursive skip connection, the same performance could be achieved by fewer parameters. }
    \label{tab:envi-transformer-fewer}
}
\end{table}

\paragraph{Parameter efficiency}

A byproduct of the proposed structure is that it could be possible to use fewer parameters to reach the same performance, especially for Transformer, which is heavily parameterized, as the expanded skip connection and layer normalization both helps to ease the optimization. Fewer parameter may translate into lower time cost.  We conduct experiments on the IWSLT-2015 EN-VI machine translation dataset using Transformer with two blocks instead of six blocks used in the main experiments. The results are shown in Table~\ref{tab:envi-transformer-fewer}. Surprisingly, a Transformer of two blocks does not perform substantially worse, and with the proposed rSkip+LN, the reduction could be easily made up. A interesting phenomenon is that the proposed method gains larger margin on deeper models, which also attests that proposed combination helps the optimization of the model, so that the performance of deeper model could be further exploited.

\paragraph{Better Optimization}

To see how the proposed combinations help the optimization of the deep neural models, in Figure~\ref{fig:grad}, we show the averaged gradient norm of the output of each residual block in pretrained ResNet-110 on CIFAR-10 based on 10,000 random training examples. Due to limited space, we only show the representative blocks at the ``middle'' of the model, meaning those blocks have the same filter size but different from the blocks lower and the blocks higher. As we can see, the original construction 1xSkip shows slight gradient ``explosion'', while simply scaling the skip causes severe gradient malformation (e.g., 2xSkip and 0.5xSkip). With normalization techniques such as layer normalization and batch normalization, the gradient malformation is generally mitigated. However, the magnitude of the gradient is different, with layer normalization more accurately reflecting the modulating factor $\lambda$ (e.g. 1xSkip and 1xSkip+LN have similar values, 2xSkip+LN is approximately twice 1xSkip, etc.), which is intuitive since batch normalization is based on mini-batch statistics unaware of the feature space. The recursive structure 2rSkip+LN is similar to 2xSkip+LN, except that its norm is larger. Especially, note the gradient at the Block 36, which receives gradient from the higher block that have different filter size; xSkip+LN and xSkip+BN seem to fail normalize its gradient properly, while rSkip+LN successfully elevates its scale, suggesting it could mitigate the difference between blocks of different nature as well. 
The last but not the least, layer normalization does not adjust the effective scale of the skip connection in forward propagation such that the effective scale becomes 1 and regresses to the plain residual structure. The learned effective scale is 2.82 and 2.64 for 2xSkip+LN and 2rSkip+LN, respectively, averaged over the shown blocks.

\begin{figure*}[t]

    \centering
    \hfil
    \includegraphics[width=0.4\linewidth]{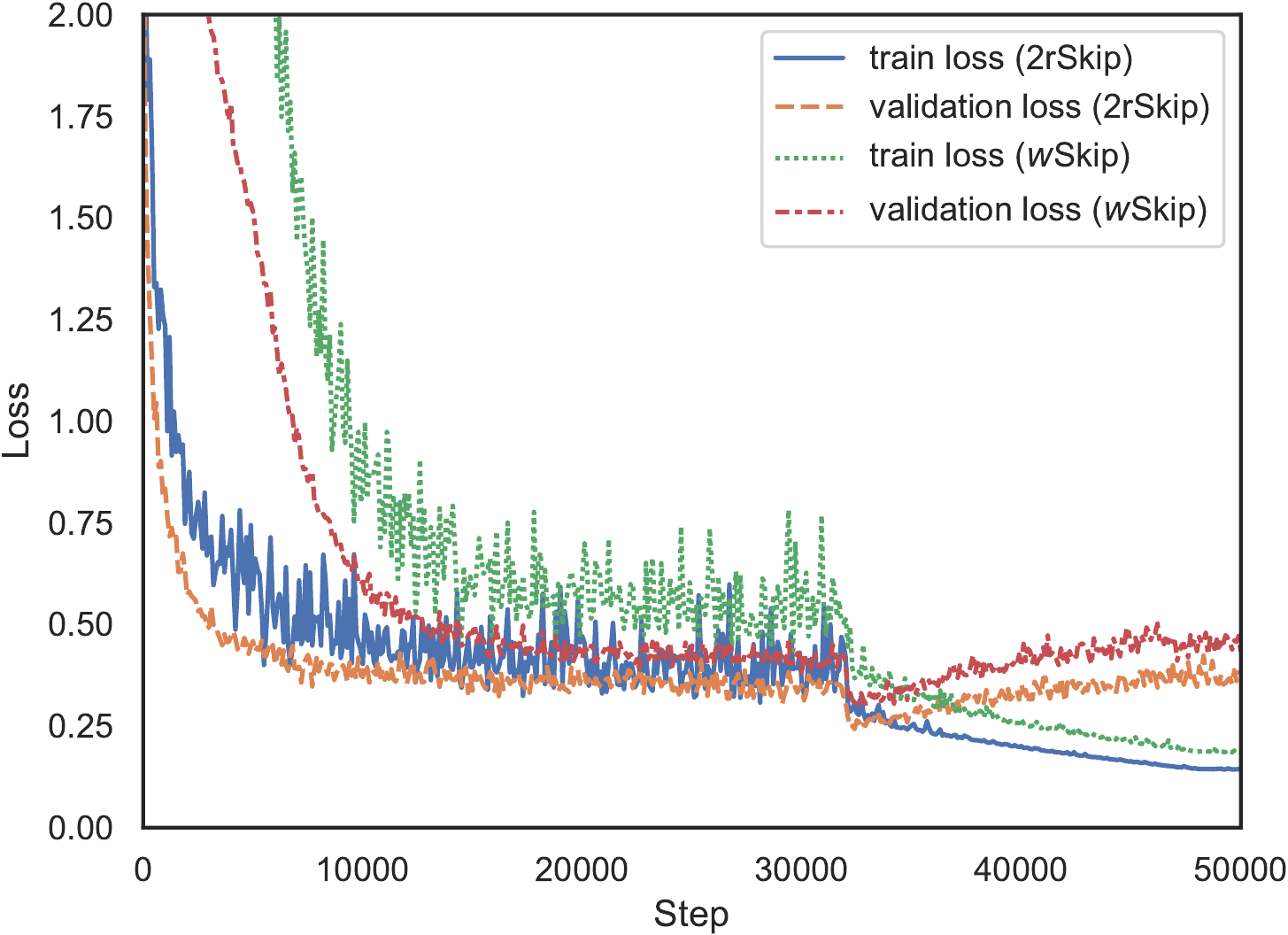}
    \hfil
    \includegraphics[width=0.4\linewidth]{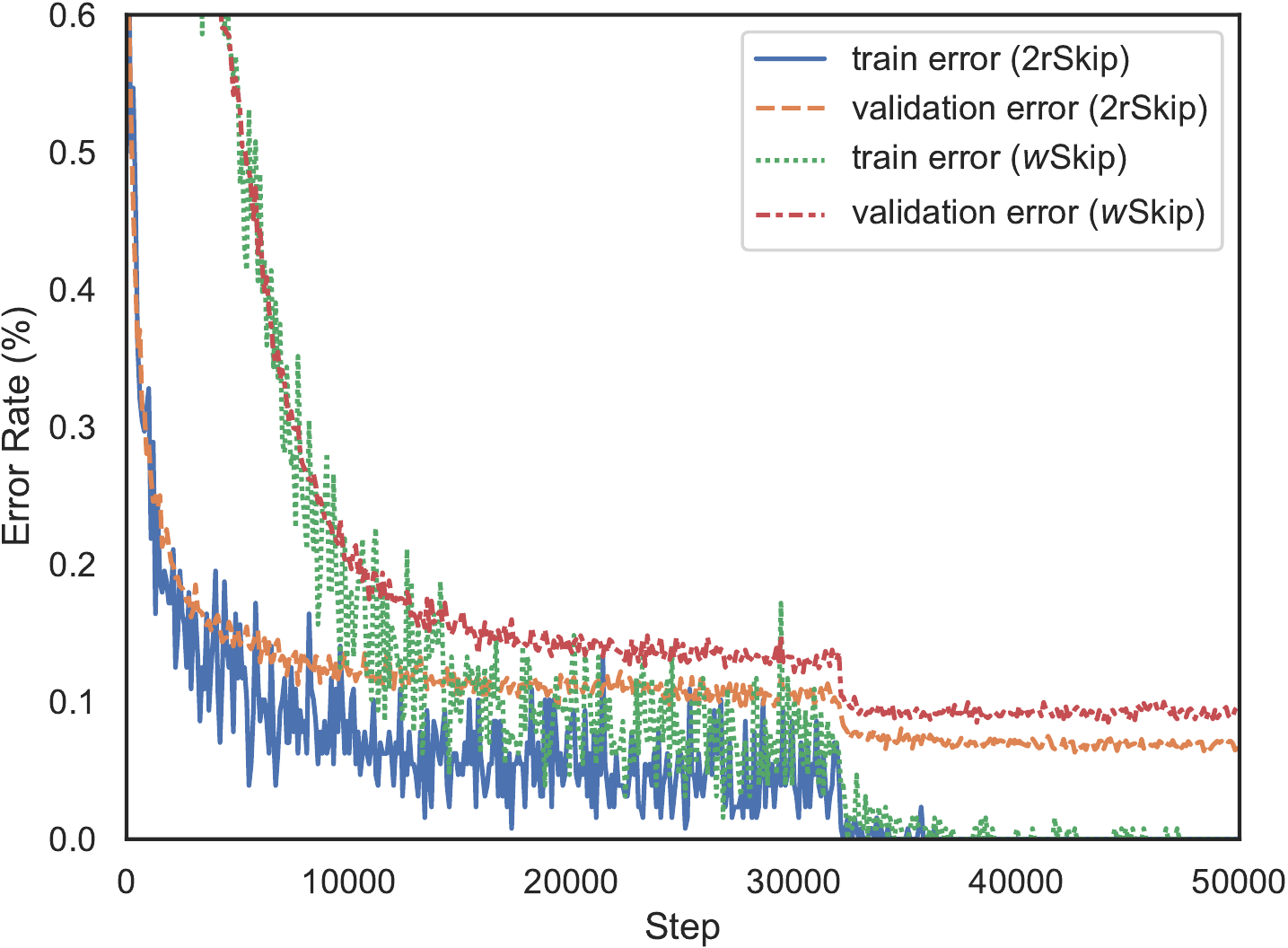}
    \hfil
    \caption{Learning curve for 2rSkip+LN and $w$Skip initialized as $\vec{1}$ on CIFAR-10 using ResNet-110. It can be observed that $w$Skip has difficulty in start training.}
    \label{fig:wx}
\end{figure*}

\paragraph{Learning $\boldsymbol{\lambda}$}

To automatically adjust the ratio between the skip and the residual branch, a trivial extension is to replace the fixed scalar $\lambda$ with a learnable scalar or vector. We also try this method on CIFAR-10 using ResNet-110. Concretely, we adopt the form of $\vec{y} = \text{LN} (\vec{w}\cdot\vec{x} + \mathcal{F}(\vec{x}))$ ($\boldsymbol{w}$\textbf{Skip+LN}), as using a parameter vector should be more general and a vector covers all the possible solutions of a scalar. However, the findings are not in favor of this approach over the fixed scalar version. When the $\vec{w}$ is initialized as $\vec{1}$, the model experiences significantly performance degradation. The averaged error rate over 5 runs is 8.66 and the final norm of a sampled $\vec{w} \in \mathbb{R}^{32}$ at middle layers is 1.34, suggesting very small scale for each dimension of the skip connection (0.24 on average). The small scale of the skip connection indicates that the model relies more on the residual branch that is parameterized and optimized for the training data, entailing a higher overfitting risk. It can be confirmed from Figure~\ref{fig:wx}, where we show the learning curve of $w$Skip compared to 2rSkip (+LN is omitted for conciseness). $w$Skip and 2rSkip both achieve low training loss, but $w$Skip has difficulty in starting learning and has higher validation loss. We also try different initialization strategy for $w$Skip, e.g., initializing $\vec{w}$ as $\vec{2}$ since we find 2 is a good empirical value for rSkip+LN. We observe a performance of 6.45 in terms of error rate, which is indeed better but still worse than rSkip+LN. Moreover, such strategy would make $w$Skip no better than using pre-determined fixed scalar values, as a proper initialization value still needs to be searched.

\subsection{Validation Results}

We verify the proposed combination of recursive skip connection with layer normalization on more difficult datasets, i.e., CIFAR-100, WMT-2014 EN-DE and EN-FR datasets. Considering the training budget, we only validate the performance of 2rSkip+LN based on the exploratory results.

The results are shown in Table~\ref{tab:c100} and Table~\ref{tab:ende}, respectively, which demonstrate the effectiveness of the proposed recursive combination that brings consistent improvements regardless of the model and the dataset. The results are in accordance with our findings in exploratory experiments as well. For CIFAR-100, the best improvement of 1.02 absolute error rates is achieved upon the baseline that is the deepest and should be the most difficult to optimize. The relative improvements across models of different depth also suggest that the representation ability of the deep models still needs to be developed and transferred into real performance elevation, while the proposal shows potential. The results on EN-VI, EN-DE and EN-FR datasets are also encouraging. With approximately one third of the parameters, the Transformer-Base model could be improved to compete with the Transformer-Big model on EN-VI and EN-DE, with the help of the refined skip connection architecture. It again demonstrates the potential of the proposal in realizing the capability of the models by inducing more efficient optimization.

\begin{table*}[t]
    \centering
    \setlength{\tabcolsep}{3pt}
    \footnotesize
    \begin{tabular}{@{}l l l l l l l@{}}
        \toprule
        Dataset     & Method    & ResNet-20 & ResNet-32 &  ResNet-44   &  ResNet-56   &  ResNet-110
        \\ \midrule
        \multirow{2}{*}{CIFAR-100} & Baseline        &  33.38  & 32.00  & 30.50  &  29.81  & 28.71\\
        & \ + proposal       &  \bf 32.99 (+0.39) & \bf 31.39 (+0.61) & \bf 29.85 (+0.65)  & \bf 28.98 (+0.83) &\bf 27.69 (+1.02) \\ 
        \bottomrule
    \end{tabular}
    \caption{Results on CIFAR-100 using ResNet and the proposed 2rSkip+LN skip connection construction. Average of 5 runs. The numbers in parentheses denote the improvements. The proposal brings consistent improvements. The improvements are more evident for deeper and complex models.}
    \label{tab:c100}
\end{table*}

\begin{table*}[t]
    \centering
    \footnotesize
    \begin{tabular}{@{}l r l l l@{}}
        \toprule
        Method   &\# Parameters & EN-VI & EN-DE& EN-FR
        \\ \midrule
        Transformer-Big \cite{Vaswani2017transformer} & 213M & - & 28.4  & 41.0 \\
        Deep Representations \cite{Dou2018Exploiting} & 356M & - & 29.2 & - \\
        Evolved Transformer \cite{So2019Evolved} & 218M & -  & \bf 29.8 & 41.3 \\
        Fixup \cite{Zhang2019Fixup}   & - & - & 29.3 & -    \\
        AdaNorm \cite{Xu2019Understanding} & - & 31.4 & 28.5 & - \\
        ScaleNorm \cite{Nguyen2019Transformers} & - & \bf 32.8 & 27.6 & -\\
        Admin  \cite{Liu2020Understanding} & - & - &  29.0 & \bf 43.8 \\
        MUSE \cite{Zhao2019MUSE}& - &  31.3 & 29.9  & 43.5 \\
        \midrule
        Transformer-Base & 65M   & 30.3  & 27.9  & 38.2  \\
        \ + proposal   & $+$0.03M &\ul{31.5 (+1.2)}& \ul{28.6 (+0.7)}&\ul{39.0 (+0.8)}  \\ \midrule
        Transformer-Big & 213M   & 31.4  & 28.5  & 41.0  \\
        \ + proposal   & $+$0.06M &\ul{32.2 (+0.8)}&\ul{29.0 (+0.5)}&\ul{41.6 (+0.6)} \\
        \bottomrule
    \end{tabular}

    \caption{Results on EN-VI, EN-DE and EN-FR using Transformer and the proposed 2rSkip+LN skip connection construction. The numbers of parameters are reported for the EN-DE models. As we can see, the proposal brings consistent substantial improvements.}
    \label{tab:ende}
\end{table*}

\section{Conclusion and Future Work}

In this work, we focus on exploring the combination of skip connection and layer normalization to build a novel skip connection architecture able to incorporate reasonable skip information at output level and solve the gradient explosion problem in practice. According to the empirical studies, we find that expanding skip information should be helpful to the optimization of the model, which is hindered by the gradient malformation in practice, and layer normalization does better than batch normalization in mitigating such problem. The proposed Recursive Skip Connection with Layer Normalization guarantees the stability of the shorcut gradient by recursively dividing the expanded skip connection and adding it to the output iteratively with layer normalization. The refined skip connection architecture may enable more efficient optimization. Validation on CIFAR-100, WMT-2014 EN-DE and EN-FR proves the findings and the effectiveness of the proposed recursive skip connection construction with best improvements over 1.0 absolute point, demonstrating its generalization ability to a wide range of existing systems.

Unfortunately, a limitation of our investigation is that $\lambda$ is selected as integer and only a small number of values are used, which may not be of the best granularity for expanding the skip connection. A more desirable solution would be to adaptively determine the value of $\lambda$, maybe in terms of the gradient distribution during the learning so that more efficient optimization could be realized. We will further seek to address this limitation in future work.

\section*{Acknowledgments}
This work is partly supported by AOTO-PKUSZ Joint Research Center for Artificial Intelligence on Scene Cognition Technology Innovation, and Beijing Academy of Artificial Intelligence (BAAI).
We thank all the anonymous reviewers for their constructive comments and suggestions. 

\bibliographystyle{coling}
\bibliography{coling2020}

\appendix

\section{On Adaptive Ratio of Recursive Skip Connection with Layer Normalization}
\label{app:ratio}
To see how the recursive structure can learn the ratio, we take $\lambda=2$ as an example and unroll the formula:
\begin{align}
    \vec{y}_2 &= \text{LN}\left(\vec{x} + \text{LN}\left(\vec{x} + \mathcal{F}(\vec{x})\right)\right) \nonumber \\
    &= \vec{w}_2 \cdot \left(\frac{  \vec{x} + \left(\vec{w}_1 \cdot \left(\frac{ \vec{x} + (\mathcal{F}(\vec{x})) - \mu_1 }{\sigma_1} \right) +  \vec{b}_1\right)  - \mu_2}{\sigma_2}\right) +  \vec{b}_2  \nonumber  \\
    &= \left(\frac{\vec{w}_2}{\sigma_2} + \frac{\vec{w}_2 \cdot \vec{w}_1}{\sigma_2 \sigma_1}\right) \vec{x}  + \frac{\vec{w}_2 \cdot \vec{w}_1}{\sigma_2 \sigma_1} \mathcal{F}(\vec{x})  + C,
\end{align}
where $\cdot$ denotes entrywise-product, $\vec{w}_1, \vec{w}_2, \vec{b}_1, \vec{b}_2$ are the gain and the bias parameters of the affine transformation in layer normalization, and $\sigma_1, \sigma_2, \mu_1, \mu_2$ are the standard deviation and the mean of $\vec{x} + \mathcal{F} (\vec{x})$ and $\vec{x} + \vec{y}_1$, respectively. As we can see, the ratio between the input and the residual is
\begin{align}
    \vec{x} : \mathcal{F}(\vec{x}) \coloneqq \frac{\vec{w}_2}{\sigma_2} + \frac{\vec{w}_2 \cdot \vec{w}_1}{\sigma_2 \sigma_1} : \frac{\vec{w}_2 \cdot \vec{w}_1}{\sigma_2 \sigma_1} = \frac{\sigma_1}{\vec{w}_1} + 1.
\end{align}
Because the recursive definition, the inner gain parameters of layer normalization could penetrate through the normalization and effectively act as the learnable ratio. It is also important to note that the ratio is also dependent on the current layer distribution, which could serve as a regularization component. For any $\lambda$ larger than 2, the general form of the ratio is $1 + \sum_{i=1}^{\lambda} \prod_{j=1}^{i} \nicefrac{\sigma_j}{\vec{w}_j}$. 
It should be reminded that setting $\lambda=2$ is sufficient for automatically learning the shortcut-residual ratio and dynamically adjusting the ratio based on instant data distribution. Yet with larger $\lambda$, the distribution of the higher order combination of $\vec{x}$ and $\mathcal{F}(\vec{x})$ will be considered. The recursive definition introduces gradual and smooth combination of the input and the residual up to the specified order $\lambda$ which could deal with different distributions better. Our experimental results also demonstrate that the proposal performs better than the scaled shortcut.

\end{document}